\documentclass[10pt,twocolumn,letterpaper]{article}

\usepackage[pagenumbers]{cvpr} 

\definecolor{cvprblue}{rgb}{0.21,0.49,0.74}
\usepackage[pagebackref,breaklinks,colorlinks,allcolors=cvprblue]{hyperref}
\usepackage{multirow}
\usepackage{amssymb}
\usepackage{pifont}
\usepackage{booktabs}
\usepackage{colortbl}  
\usepackage{xcolor}
\usepackage{makecell,pifont,multirow,multicol}
\usepackage{colortbl}
\usepackage{caption}
\usepackage{subcaption}
\usepackage{amssymb}
\usepackage{marvosym}
\usepackage{titletoc}
\usepackage{color}
\usepackage{multicol}
\usepackage{pifont}
\usepackage{array}
\usepackage[table]{xcolor} 
\usepackage{pifont}        
\usepackage{graphicx}      

\usepackage[accsupp]{axessibility} 
\usepackage{mathtools,bm}
\usepackage{microtype}

\def\paperID{N/A} 
\def\confName{CVPR}
\def\confYear{2026}

\title{ColaVLA: Leveraging Cognitive Latent Reasoning for Hierarchical Parallel Trajectory Planning in Autonomous Driving}

\author{Qihang Peng\textsuperscript{1,2,3} \quad
Xuesong Chen\textsuperscript{2, 3, $\dagger$}  \quad
Chenye Yang\textsuperscript{1} \quad
Shaoshuai Shi\textsuperscript{3, \Letter}   \quad
Hongsheng Li\textsuperscript{2, \Letter} \\
\protect \textsuperscript{1} Tsinghua University \quad
\protect \textsuperscript{2} CUHK MMLab \quad
\protect \textsuperscript{3} Voyager Research, Didi Chuxing \\
{\tt\small  pqh22@mails.tsinghua.edu.cn, shaoshuaics@gmail.com, hsli@ee.cuhk.edu.hk} \\
{\small $^\dagger$Project Leader\hspace{0.4cm}  \textsuperscript{\Letter}Corresponding author}
}

\begin{document}
\maketitle
\begin{abstract}

Autonomous driving requires generating safe and reliable trajectories from complex multimodal inputs. Traditional modular pipelines separate perception, prediction, and planning, while recent end-to-end (E2E) systems learn them jointly. Vision–language models (VLMs) further enrich this paradigm by introducing cross-modal priors and commonsense reasoning, yet current VLM-based planners face three key challenges: (i) a mismatch between discrete text reasoning and continuous control, (ii) high latency from autoregressive chain-of-thought decoding, and (iii) inefficient or non-causal planners that limit real-time deployment. We propose \textbf{ColaVLA}, a unified vision–language–action framework that transfers reasoning from text to a unified latent space and couples it with a hierarchical, parallel trajectory decoder. The \textbf{Cognitive Latent Reasoner} compresses scene understanding into compact, decision-oriented meta-action embeddings through ego-adaptive selection and only two VLM forward passes. The \textbf{Hierarchical Parallel Planner} then generates multi-scale, causality-consistent trajectories in a single forward pass. Together, these components preserve the generalization and interpretability of VLMs while enabling efficient, accurate and safe trajectory generation. Experiments on the nuScenes benchmark show that ColaVLA achieves state-of-the-art performance in both open-loop and closed-loop settings with favorable efficiency and robustness.
The project is at \url{https://github.com/pqh22/ColaVLA}.
\end{abstract}
    
    \label{sec:intro}
\section{Introduction}


\begin{figure}[t]
\centering
\includegraphics[width=0.45\textwidth ]{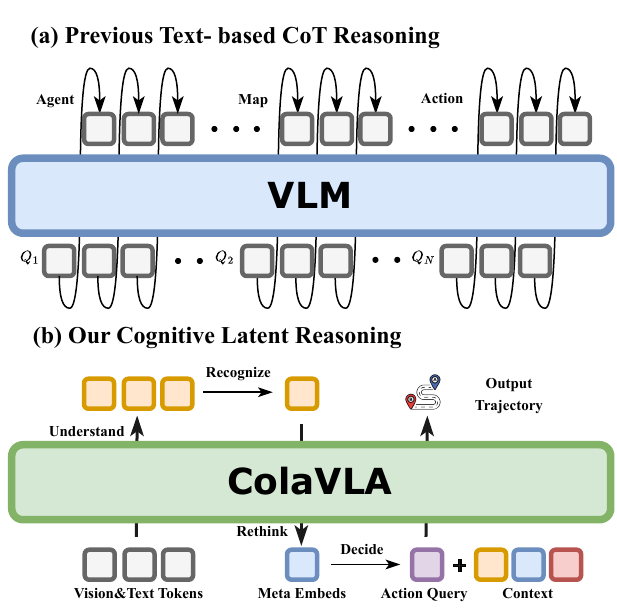}
\vspace{-1mm}
\caption{\textbf{Illustration of inference paradigms.} (a) Previous driving VLMs use text-based chain-of-thought, autoregressively emitting intermediate texts for sub-tasks; repeated decoding increases token cost and error compounding, causing high latency. (b) Our model performs latent reasoning in a VLA space with \textbf{three} forward passes, \ie scene understanding, latent rethink, and parallel action decoding, removing autoregressive text and cutting inference latency while preserving decision-level interpretability.}

\vspace{-3mm}
\label{fig:illustrate}
\end{figure}

Autonomous driving aims to predict safe and comfortable motion from rich multimodal observations. Early systems were organised as modular perception–prediction–planning stacks with dedicated 3D perception~\cite{qi2017pointnet,lang2019pointpillars,sun2020scalability,li2022bevformer, zheng2024denseg, zheng2025densegrounding, peng2025proxytransformation}, forecasting~\cite{chai2019multipath,gu2023vip3d,liu2021multimodal,gu2021densetnt}, and planning components~\cite{song2020pip,ettinger2021large,caesar2021nuplan,chitta2022transfuser,liao2024diffusiondrive}. Recent end-to-end approaches unify the stack and learn from pixels to waypoints or controls in a single pipeline~\cite{hu2022stp3,jiang2023vad,hu2023planning,chen2024vadv2,zheng2024genad,weng2024drive,sun2024sparsedrive}. In parallel, vision–language models (VLMs)~\cite{gpt3,DeepSeek-VL2,bai2025qwen2,wang2023cogvlm,liu2024visual} are increasingly integrated to inject cross-modal priors and world knowledge, either as agentic planning models that output controls or trajectories~\cite{xu2024drivegpt4,mao2023language,sima2023drivelm,hwang2024emma,xing2024openemma,wang2024omnidrive} or as reasoning assistants that guide end-to-end modules~\cite{jiang2024senna,tian2024drivevlm,Chen_2025_CVPR}.

Despite rapid progress, important gaps remain between accuracy in benchmarks and reliability in deployment. Modular systems provide interpretable components and strong geometric priors, but brittle interfaces can propagate errors and make global optimization difficult. 
End-to-end systems reduce manual interfaces and achieve high open-loop accuracy~\cite{hu2022stp3,jiang2023vad,hu2023planning,chen2024vadv2,zheng2024genad,weng2024drive}, yet they often rely on sparse trajectory supervision, intertwine perception and control in ways that obscure causal structure, and struggle to generalize to out-of-distribution scenarios.
Text-based VLM planners add powerful priors but introduce some practical issues: (1) Modality mismatch. Discrete text tokens do not align with the continuous geometry and dynamics of trajectories and can lead to format violations or physically inconsistent waypoints~\cite{mao2023language,hwang2024emma}. (2) Chain-of-thought reasoning latency. The computation overhead of autoregressive decoding arises from its iterative token-by-token generation, where each new token depends on previously ones, causing the sequence to grow over time and significantly increasing inference delay. 

To bridge the above gaps, we revisit the role of the VLM in the driving task and introduce a transition from explicit textual chains of thoughts to unified latent reasoning. Our key idea is to execute reasoning entirely in a unified latent space and to pair it with a planner that preserves causal structure while decoding in parallel. This retains the knowledge priors and reasoning ability of VLMs, yet avoids lengthy autoregressive reasoning and its latency. At a high level, ColaVLA consists of a \textbf{Cognitive Latent Reasoner} that distills scene evidence into compact meta-action priors, and a \textbf{Hierarchical Parallel Planner} that converts these priors to multi-scale trajectories under a causality-preserving scheme.

First, the cognitive latent reasoner efficiently accomplishes scene understanding and the final meta-action decision through two forward passes. Specifically, during the first forward pass, the reasoner constructs a multimodal input sequence comprising a fixed driving prompt, multi-view visual images, and ego status, and passes it through a VLM to obtain unified tokens that have undergone complete contextual interaction. However, the visual tokens contain substantial redundant information irrelevant to the driving decision. To extract decision-relevant information from the scene, we introduce an ego-adaptive modulation to align these tokens with the instantaneous vehicle state, after which a lightweight router scores and selects the top-$K$ safety-critical vision tokens. Subsequently, for the second forward pass, we concatenate the selected context with learnable meta queries as input, enabling each meta-action embedding to query the driving-critical context via cross-attention and ultimately yield the final driving decision. Guided by the reasoner's decision, the hierarchical parallel planner, using the same VLMs as our reasoner, predicts multi-scale fine-grained trajectories through parallel decoding. Specifically, we retrieve the corresponding meta-action embedding from an action bank based on the chosen decision. With the selected meta-action, we instantiate a full-time-domain action block using temporal embeddings, which is then resampled into $S$ nested, coarse-to-fine scales. Finally, the embeddings from all scales are concatenated with the pruned context to serve as the planner's input, which are then decoded into trajectories in parallel within a single forward pass. This design yields a coherent and causally consistent planning process and produces multi-scale continuous trajectories with substantially reduced inference latency.

Our main contributions are as follows. (1) We present ColaVLA, a unified vision–language–action framework for end-to-end autonomous driving that operates directly on continuous trajectories, avoiding modality mismatch while leveraging VLM priors. (2) We design a Cognitive Latent Reasoner that relocates reasoning from textual chain-of-thought to a unified latent space, allowing the model to observe broadly, focus selectively, rethink carefully and decide efficiently through ego-adaptive routing and meta information compression. (3) We propose a Hierarchical Parallel Planner that decodes all temporal scales and modes in a single forward pass, achieving efficient, reasonable and safe trajectory generation under tight latency constraints. (4) Comprehensive experiments on the nuScenes benchmark demonstrate that ColaVLA establishes new state-of-the-art performance in both open-loop and closed-loop evaluations, while maintaining strong interpretability and computational efficiency.

\section{Related Work}
\label{sec:related_work}

\subsection{End-to-end Autonomous Driving}
Classical autonomous driving stacks decompose perception, prediction, and planning into separate modules, easing engineering but causing information loss and suboptimal coordination. 
End-to-end (E2E) systems instead learn a single differentiable mapping from sensor inputs to trajectories or control commands. 
Early demonstrations proved feasibility but suffered from limited transparency and data efficiency. 
Recent works enhance this paradigm with richer supervision and unified architectures: UniAD~\cite{hu2023planning} integrates perception and planning within a Transformer, and VAD~\cite{jiang2023vad,chen2024vadv2} employs vectorized abstractions for efficiency and uncertainty modeling. 
Other directions explore task dependency~\cite{weng2024drive}, latent generative modeling~\cite{zheng2024genad,liao2024diffusiondrive}, or simplified ego-centric controllers~\cite{admlp,li2024ego}. 
Despite progress, most E2E methods rely on sparse supervision from a single ground truth trajectory, which limits reasoning and diversity, highlighting the need for higher-level semantic and cognitive guidance.

\subsection{VLMs for Autonomous Driving}
Vision–Language Models (VLMs) bring broad world knowledge and remarkable reasoning ability to autonomous driving, inspiring two main integration paradigms. 
\textit{Single-system} approaches treat planning as language modeling: visual inputs and prompts are translated into textual trajectories or control commands with chain-of-thought explanations~\cite{hwang2024emma,mao2023gpt,xing2024openemma,xu2024drivegpt4,chen2024driving}. 
\textit{Dual-system} designs pair a VLM with an end-to-end planner, where the VLM proposes high-level commands or low-frequency trajectories to guide a lightweight action controller~\cite{tian2024drivevlm,jiang2024senna,Chen_2025_CVPR}. 
Other efforts align vision–language features in BEV~\cite{wang2024drive}, extend reasoning to 3D~\cite{wang2024omnidrive}, or use VLMs as teachers to supervise E2E planning models~\cite{xu2024vlm}. 
While these methods improve interpretability and generalization, they still rely on explicit text-level reasoning or suffer from a feature gap between the VLM and the downstream action planner. This motivates our unified vision–language–action framework, which performs reasoning entirely in a shared latent space, bridging perception, cognition, and control within a single coherent architecture.

\begin{figure*}[t]
    \centering
    \includegraphics[width=\linewidth]{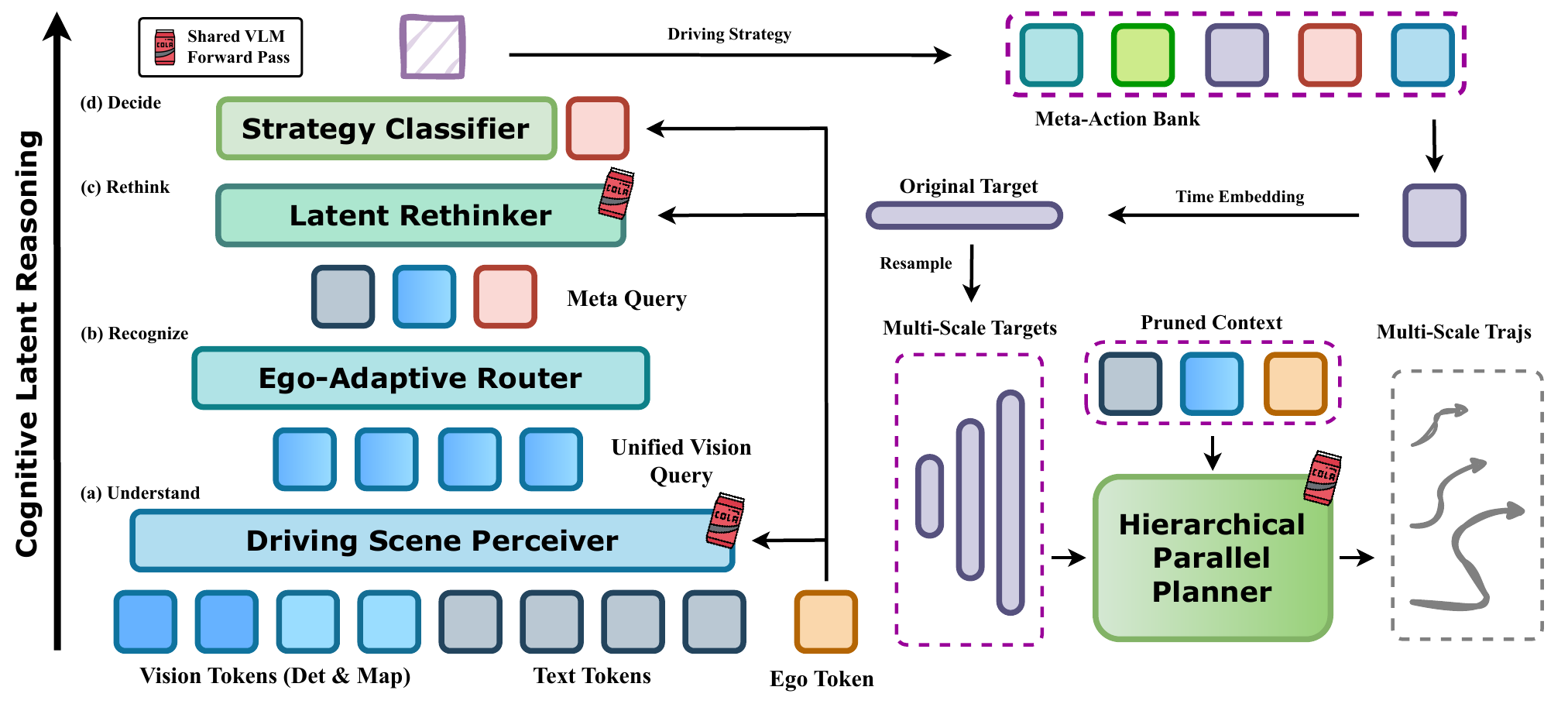}
    \caption{
    Overview of the \textbf{ColaVLA} framework.
    Multi-view image sequences are first processed by an image backbone and a Q-Former to perceive 3D objects and vectorized maps, producing visual tokens for subsequent reasoning and planning. 
    On the left, the \textbf{Cognitive Latent Reasoning} module performs implicit reasoning through four stages, \ie \textit{Understand}, \textit{Recognize}, \textit{Rethink}, and \textit{Decide}, to derive a driving strategy. 
    On the right, the derived strategy selects corresponding meta-action queries from action bank, which are then transformed to multi-scale targets. These targets, together with the pruned context are fed into a \textbf{Hierarchical Parallel Planner} for one-pass, parallel trajectory decoding. 
    The resulting multi-scale trajectories enable efficient, causal, and interpretable end-to-end planning.
    }
    \label{fig:framework}
\end{figure*}

\section{Method}
\label{sec:method}
\subsection{Framework Overview}
\label{sec:formulation}

Given multimodal representation 
$\mathbf{S}_t$
comprising multi–view images, LiDAR points, radar grids, a text prompt and ego state, or a subset thereof,
the trajectory planning task reduces to mapping $\mathbf{S}_t$ to a $K$-step trajectory
$\widehat{\mathbf{Y}}_t=[(x_{t+1},y_{t+1}),\ldots,(x_{t+K},y_{t+K})]$. Existing planning systems generally consist of two unified learnable components:
\emph{\textsc{\textbf{Reasoner}}} ($\mathcal{R}_{\theta}$), usually implemented as ResNet~\cite{he2016resnet}, PointNet~\cite{qi2017pointnet}, Q-Former~\cite{dalmonte2025qformerautoencodermodernframework}, or VLM Transformer that extracts and fuses multimodal features or optionally refines them with chain-of-thought, and
\emph{\textsc{\textbf{Planner}}} ($\mathcal{P}_{\phi}$), usually instantiated as an deterministic MLP planner, stochastic diffusion sampler, or standard Transformer~\cite{attention} decoder 
that injects action queries and regresses continuous way-points.
Their interaction is succinctly captured by
\begin{equation}
\mathbf{Z}_t = \mathcal{R}_{\theta}(\mathbf{S}_t)
\qquad
\widehat{\mathbf{Y}}_t = \mathcal{P}_{\phi}(\mathbf{Z}_t,\mathbf{A}),
\end{equation}
where $\mathbf{Z}_t\!\in\!\mathbb{R}^{L\times D}$ is a latent token sequence and
$\mathbf{A}$ represents a learnable action bank.
This abstraction subsumes three dominant paradigms in the field: (1) classical modular pipelines with separate perception, decision, and planning stages; (2) end-to-end planners without explicit reasoning; and (3) VLM-based models that formulate planning as autoregressive text generation with chain-of-thought reasoning.

To unify the generalization ability of VLMs with the efficiency of action-based planners, we reformulate trajectory planning under a unified VLA framework as shown in~\cref{fig:framework}. 
Rather than relying on textual chain-of-thought reasoning, we introduce \textbf{Cognitive Latent Reasoning} in~\cref{sub:reasoning} that operates fully in the unified latent space, requiring only two forward passes to yield an interpretable, decision-oriented meta representation. 
This compact meta-action prior captures high-level intent and contextual awareness without the substantial latency of autoregressive decoding. 
Built upon this, ColaVLA employs a \textbf{Hierarchical Parallel Planner} (detailed in~\cref{sub:planning})  that performs structured trajectory refinement under a causality-preserving hybrid attention mask. 
This design enables multi-scale trajectory generation and parallel multi-mode decoding in a single forward pass, bridging knowledge-driven reasoning and continuous control for a unified, interpretable, and low-latency planning pipeline.

\subsection{Cognitive Latent Reasoning}
\label{sub:reasoning}
Recent vision–language models show strong generalization in autonomous driving by performing a driver-like reasoning process: perceiving the scene, identifying critical entities, rethinking carefully and deciding final strategies. 
However, most existing approaches realize this reasoning at the text level, leading to high latency and exposure bias from autoregressive decoding. 
Inspired by advances in latent-space reasoning for large language models~\cite{hao2024training}, we shift this cognitive process to a unified latent space. Our reasoner performs that within only two forward passes, preserving the semantic richness of language-based reasoning while avoiding token-generation overhead. This interpretable design forms an efficient foundation for hierarchical parallel planning.

\noindent\textbf{Driving Scene Comprehension.}\;
The cognitive process of a human driver begins with a holistic survey
of the environment.  
To emulate this step, we form an input sequence by mixing a
fixed driving prompt encoded as text embeddings
$ \mathbf T\!\in\!\mathbb{R}^{ L_t\times D}$,  
multi-view vision embeddings
$ \mathbf V\!\in\!\mathbb{R}^{ L_v\times D}$ produced by the
perception front-end,  
and an ego token
$\mathbf E\!\in\!\mathbb{R}^{1\times D}$ that represents the
vehicle state.  
The shared VLM transformer $\mathcal D_{\textsc{vlm}}$ processes this
sequence, and we keep only the visual slice of the hidden states:
\begin{equation}
\mathbf Q_{\textsc V}
  =
     \mathcal D_{\textsc{vlm}}
           ([\ \mathbf T; \mathbf V; \mathbf E\ ])
  \in\mathbb{R}^{ L_v\times D}.
\label{eq:first_pass}
\end{equation}
The updated text and ego embeddings are discarded so that the prompt
remains immutable and no redundant information is introduced.
And $\mathbf Q_{\textsc V}$ thus provides a temporally
causal, globally coherent representation of spatial semantics, lane
topology, and dynamic agents, establishing a stable foundation for the
subsequent identification of critical entities.

\noindent\textbf{Critical Entity Recognition.}
Selecting only safety-critical cues is essential for both efficiency and reliability. 
We introduce an ego-adaptive router $\mathcal H_{\!\phi}$ that first applies FiLM conditioning to align visual tokens with the current ego state:
\begin{equation}
\tilde{\mathbf Q}_{\textsc V}
  =\bigl(1+\gamma_{\textsc Re}(\mathbf E)\bigr)\odot\mathbf Q_{\textsc V}
   +\beta_{\textsc Re}(\mathbf E) \in\mathbb{R}^{L_v\times D},
\label{eq:film_router}
\end{equation}
where the scale $\gamma_{\textsc Re}(\mathbf E)$ and shift $\beta_{\textsc Re}(\mathbf E)$ are generated from the ego token through two independent linear projections. 
This conditioning highlights scene elements consistent with the ego vehicle’s velocity, heading, and curvature, emphasizing dynamic agents and lane boundaries within the collision cone while suppressing irrelevant background details.

Next, the router evaluates the ego-modulated tokens and then selects the most informative subset:
\begin{equation}
\mathbf w = \mathcal H_{\phi}\bigl(\tilde{\mathbf Q}_{\textsc V}\bigr)\in\mathbb R^{L_v},\
\mathbf Q^{*} = \operatorname{TopK}\bigl(\tilde{\mathbf Q}_{\textsc V},\, \mathbf w,\, K\bigr).
\label{eq:router_select}
\end{equation}
During training, the selection is made differentiable using a Gumbel–Softmax relaxation to form a $K$-hot mask, while at inference time the top-$K$ tokens are retained directly. 
The resulting compact set $\mathbf Q^{*} \in \mathbb R^{K \times D}$ effectively captures the most safety-relevant visual cues, \eg lanes, nearby vehicles, pedestrians, and traffic lights, serving as an efficient information bottleneck for the subsequent latent reasoning stage.

\noindent\textbf{Latent Rethinking.} After pruning irrelevant cues, a human driver would reassess the condensed evidence and form a provisional plan. 
We emulate this process by concatenating four components: the fixed driving prompt $\mathbf T$, the $K$ salient vision tokens $\mathbf Q^{*}\!\in\!\mathbb{R}^{K\times D}$, the ego token $\mathbf E\in\mathbb{R}^{1\times D}$, and a bank of $C$ learnable meta-queries $\mathbf M=[\mathbf m_1,\dots,\mathbf m_C]\!\in\!\mathbb{R}^{C\times D}$. 
This sequence is passed through a second forward pass of the shared VLM transformer:
\begin{equation}
\mathbf Q_{\textsc M}
  = \mathcal D_{\textsc{vlm}}
      \bigl([\ \mathbf T;\mathbf Q^{*};\mathbf E;\mathbf M\ ]\bigr)
  \in \mathbb{R}^{C\times D}.
\label{eq:second_pass}
\end{equation}
Sharing VLM ensures temporal and semantic consistency, while the smaller token budget ($C\!\ll\!L_v$) keeps computation efficient. 
Each $\mathbf m_c$ is initialized to represent a corresponding meta-action, such as straight cruising, unprotected left turn, or hard braking, derived from clustering training trajectories. The updated embeddings $\mathbf Q_{\textsc M}$ represents different driving strategies, which are ready for final decision-making.

\noindent\textbf{Strategic Decision Synthesis.} The meta-query embeddings $\mathbf Q_{\textsc M}$ are adapted to the ego state through FiLM modulation and cross-attend to the driving-critical visual tokens $\mathbf Q^{*}$, followed by a self-attention layer over the meta tokens. 
A shared two-layer MLP then maps each refined meta token to a maneuver logit, trained with a focal loss that emphasizes hard and safety-critical cases. 
By constraining the reasoning space to $C$ meta-action tokens, the process achieves entropy reduction and produces multiple probable driving strategies, which form structured priors for subsequent prediction.

\subsection{Hierarchical Parallel Planning}
\label{sub:planning}

We introduce a Hierarchical Parallel Planner that generates trajectories through multi-stage intent-to-motion decoding. 
It integrates temporal abstraction, a causality-preserving attention scheme, and confidence-guided multi-mode decoding, with three key properties:  
(i) intent-to-motion refinement, where final destination is progressively resolved into detailed motion plans;
(ii) stage-wise independence, enabling parallel decoding across temporal scales for efficient inference; and (iii) confidence-aware diversity, maintaining multiple plausible modes to enhance robustness.

\noindent \textbf{Stage-Aware Trajectory Querying.}
Given a prediction horizon of $T$ steps, $\mathcal{T}\!=\!\{0,\dots,T{-}1\}$, we divide it into $S$ nested stages $\mathcal{I}_1 \subset \dots \subset \mathcal{I}_S = \mathcal T$, where earlier stages represent coarse temporal resolutions and later stages progressively refine the trajectory. 
For each scale $s$, the meta-action query $\mathbf{A} \in \mathbb R^D$ selected by the cognitive reasoner is expanded with temporal embeddings to form trajectory targets $\mathbf{F}\!\in\!\mathbb R^{T\times D}$, which are resampled into multi-scale subsets $\mathbf{F}_s \in \mathbb R^{|\mathcal I_s| \times D}$ following the predefined order. 
The pruned context and all scale-wise targets are then concatenated in temporal order to form the complete multi-scale input stream:
\begin{equation}
\mathbf{X} = [\mathbf{Q}^*;\mathbf{F}_1;\dots;\mathbf{F}_S] \in \mathbb R^{L \times D},\ L = K+\sum_{s=1}^{S}|\mathcal I_s|
\label{eq:querying}
\end{equation}
where $\mathbf{Q}^*$ represents the pruned context encoding the historical and environmental information. 
This hierarchical composition ensures that coarse trajectories precede finer refinements, establishing the foundation for the subsequent causality-preserving masking mechanism to perform structured multi-mode planning within a single forward pass.

\begin{figure}[t]
\centering
\includegraphics[width=0.48\textwidth]{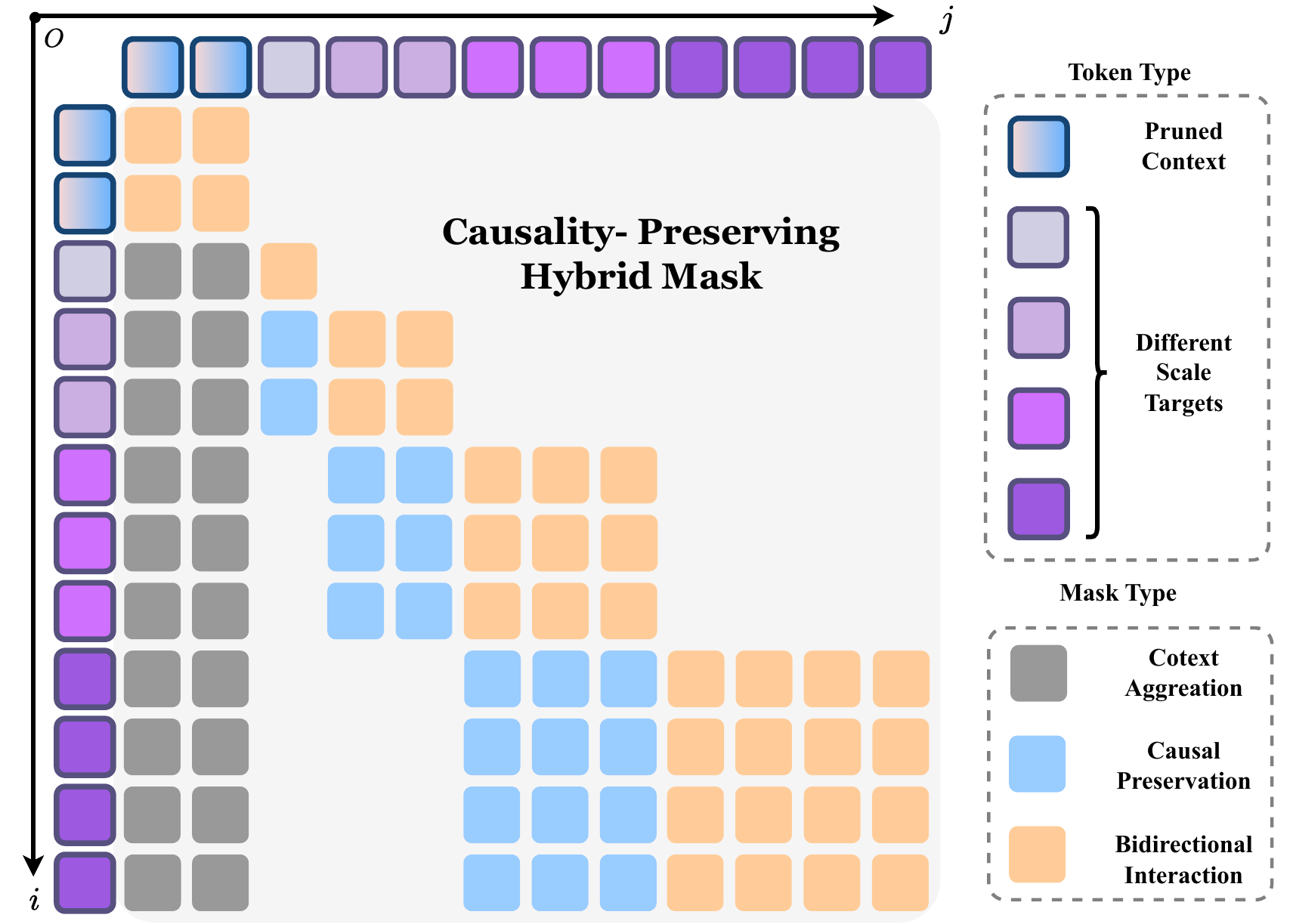}
\caption{
\textbf{Causality-Preserving Hybrid Mask.}
Our mask is designed for the multi-scale targets within our planner. 
It enables information flow from the pruned context to all temporal scales, while maintaining temporal causality between adjacent scales. 
}
\label{fig:mask}
\end{figure}

\noindent \textbf{Causality-Preserving Hybrid Attention.}
We design a hybrid attention mask $\mathcal{M}$ to regulate information flow between the pruned context and multi-scale trajectory tokens as shown in \cref{fig:mask}. 
It satisfies three principles: (i) bidirectional interaction within each category, enabling context tokens and same-scale trajectory tokens to maintain local consistency; (ii) global context aggregation, allowing every trajectory token to attend to all context tokens; and (iii) causal preservation, where tokens at scale $s$ can only access those from the preceding coarser scale $s{-}1$, preventing information leakage from future finer scales.

Formally, the mask is defined as
\begin{equation}
\mathcal{M}(i,j)=
\begin{cases}
  0, & j \le L_c, \\[4pt]
  0, & i \ge L_c \ \text{and}\ \mathbf{X}[j] \in \mathcal{I}_{s-1} \cup \mathcal{I}_{s}, \\[4pt]
  -\infty, & \text{otherwise},
\end{cases}
\label{eq:hybrid_mask}
\end{equation}
where $L_c$ is the length of pruned context and $s$ indexes the current scale. $\mathbf{X}$ denotes the concatenated sequence. 
The mask thus allows each token at scale $s$ to attend to both the pruned context and the immediately preceding scale $\mathcal{I}_{s-1}$, while preventing access to future scales $\mathcal{I}_{s+1}, \dots, \mathcal{I}_{S}$. 
This design effectively combines global context aggregation with strictly causal refinement, ensuring that trajectory decoding proceeds in a physically consistent coarse-to-fine manner.

\noindent \textbf{Confidence-Guided Parallel Decoding.}
In the final decoding stage, the planner simultaneously processes multiple candidate driving strategies, each producing a set of latent trajectory targets. 
Two lightweight MLP heads are applied independently to estimate the confidence score and to regress the corresponding multi-scale trajectory for each hypothesis. 
During training, a one-hot supervision signal is assigned based on the distance between each predicted trajectory and the ground-truth, where only the closest hypothesis receives direct regression supervision. 
This confidence-guided mechanism enables the model to prioritize the most reliable trajectory while preserving diversity across hypotheses. 
The parallel decoding further processes all candidates within a single forward pass, ensuring high efficiency and enhanced generalization by effectively preventing mode collapse.

\section{Experiments}

\begin{table*}[ht!]
\centering
{\begin{tabular}[b]{l|c|c|cccc|cccc}
\toprule[1.5pt]
\multirow{2}{*}{\textbf{Method}} &
\multirow{2}{*}{\textbf{Reference}} &
\multirow{2}{*}{\textbf{Ego}} &
\multicolumn{4}{c|}{\textbf{L2 ($m$)} $\downarrow$} &
\multicolumn{4}{c}{\textbf{Col. Rate (\%)} $\downarrow$ }\\
& & & 1$s$ & 2$s$ & 3$s$ & Avg. & 1$s$ & 2$s$ & 3$s$ & Avg. \\
\midrule

\multicolumn{11}{c}{\textit{Text-Based Driving Models}} \\
\midrule
DriveVLM~\cite{tian2024drivevlm} & CoRL 2024 & \ding{51} & 0.18 & 0.34 & 0.68 & 0.40 & -- & -- & -- & -- \\
DriveVLM-Dual~\cite{tian2024drivevlm} & CoRL 2024 & \ding{51} & 0.15 & 0.29 & 0.48 & 0.31 & -- & -- & -- & -- \\
OmniDrive~\cite{wang2024omnidrive} & CVPR 2025 & \ding{51} & 0.14 & 0.29 & 0.55 & 0.33 & \textbf{0.00} & \textbf{0.13} & 0.78 & 0.30 \\
EMMA~\cite{hwang2024emma} & TMLR & \ding{51} & 0.14 & 0.29 & 0.54 & 0.32 & -- & -- & -- & -- \\
EMMA+~\cite{hwang2024emma} & TMLR & \ding{51} & \textbf{0.13} & 0.27 & 0.48 & 0.29 & -- & -- & -- & -- \\
ImpromptuVLA~\cite{chi2025impromptu} & NeurIPS 2025  &\ding{51} &\textbf{0.13}          & 0.27      & 0.53       &  0.30      & -- & -- & --  &  -- \\
SOLVE-VLM~\cite{Chen_2025_CVPR} & CVPR 2025 & \ding{51} & \textbf{0.13} & \textbf{0.25} & \textbf{0.47} & \textbf{0.28} & \textbf{0.00} & 0.16 & 0.43 & \textbf{0.20} \\
\midrule

\multicolumn{11}{c}{\textit{Action-Based Driving Models}} \\
\midrule
UniAD~\cite{hu2023planning} & CVPR 2023 & -- & 0.59 & 1.01 & 1.48 & 1.03 & 0.16 & 0.51 & 1.64 & 0.77 \\
VAD-Base~\cite{jiang2023vad} & ICCV 2023 & -- & 0.69 & 1.22 & 1.83 & 1.25 & 0.06 & 0.68 & 2.52 & 1.09 \\
BEV-Planner~\cite{li2024ego} & CVPR 2024 & -- & 0.30 & 0.52 & 0.83 & 0.55 & 0.10 & 0.37 & 1.30 & 0.59 \\
UniAD~\cite{hu2023planning} & CVPR 2023 & \ding{51} & 0.20 & 0.42 & 0.75 & 0.46 & 0.02 & 0.25 & 0.84 & 0.37 \\
VAD-Base~\cite{jiang2023vad} & ICCV 2023 & \ding{51} & 0.17 & 0.34 & 0.60 & 0.37 & 0.04 & 0.27 & 0.67 & 0.33 \\
AD-MLP~\cite{admlp} & arXiv 2023 & \ding{51} & 0.15 & 0.32 & 0.59 & 0.35 & \textbf{0.00} & 0.27 & 0.85 & 0.37 \\
BEV-Planner++~\cite{li2024ego} & CVPR 2024 & \ding{51} & 0.16 & 0.32 & 0.57 & 0.35 & \textbf{0.00} & 0.29 & 0.73 & 0.34 \\
SOLVE-E2E~\cite{Chen_2025_CVPR} & CVPR 2025 & \ding{51} & 0.14 & 0.28 & 0.50 & 0.31 & 0.04 & \textbf{0.17} & 0.68 & 0.30 \\
 \rowcolor{gray!20}
ColaVLA & -- & \ding{51} & \textbf{0.14} & \textbf{0.27} & \textbf{0.50} & \textbf{0.30} & 0.04 & \textbf{0.17} & \textbf{0.47} & \textbf{0.23} \\
\bottomrule[1.5pt]
\end{tabular}}
\caption{Open-loop planning results on the nuScenes benchmark. Methods are grouped into text-based driving models (top) and action-based driving models (bottom). Within action-based approaches, ColaVLA achieves the best overall results, \ie lowest average L2 and the best collision rates, demonstrating accuracy and safety while retaining high inference efficiency.}

\label{tab:sota-plan}
\end{table*}

\begin{table*}[t]
\centering
\footnotesize
\resizebox{0.9\linewidth}{!}{%
\begin{tabular}{lc@{\hspace{1.5em}}cccc@{\hspace{1.5em}}cccc} 
\toprule[1.5pt]
\multirow{2}{*}{\textbf{Method}} & \multirow{2}{*}{\textbf{Reference}} & \multicolumn{4}{c}{\textbf{NeuroNCAP Score $\uparrow$}} & \multicolumn{4}{c}{\textbf{Collision rate (\%) $\downarrow$}} \\
\cmidrule(lr){3-6} \cmidrule(lr){7-10} 
& & \textbf{Avg.} & \textbf{Stat.} & \textbf{Frontal} & \textbf{Side} & \textbf{Avg.} & \textbf{Stat.} & \textbf{Frontal} & \textbf{Side} \\ 
\midrule
 UniAD~\cite{hu2023planning}  &   CVPR 2023    & 0.73          & 0.84   & 0.10    & 1.26   & 88.6          & 87.8   & 98.4   & 79.6   \\
 VAD~\cite{jiang2023vad} & ICCV 2023         & 0.66          & 0.47   & 0.04    & 1.45   & 92.5          & 96.2   & 99.6   & 81.6   \\
 SparseDrive~\cite{sun2024sparsedrive} &ICRA 2025 & 0.92          & --      & --       & --      & 93.9          & --      & --      & --      \\
 BridgeAD-S~\cite{zhang2025bridging} & CVPR 2025 &1.52          & --      & --       & --      & 76.2          & --      & --      & --      \\

 BridgeAD-B~\cite{zhang2025bridging} & CVPR 2025 &1.60          & --      & --       & --      & 72.6          & --      & --      & --      \\
ImpromptuVLA$^\dagger$~\cite{chi2025impromptu} & NeurIPS 2025 &2.06          & 2.55      & 1.86       & 1.78      & 65.1 & 54.8 & 72.8  &  67.6  \\
BridgeAD-S$^\ddagger$~\cite{zhang2025bridging} & CVPR 2025 &2.98          & --      & --       & --      & 46.1          & --      & --      & --      \\
 BridgeAD-B$^\ddagger$~\cite{zhang2025bridging} & CVPR 2025 &3.06          & --      & --       & --      & 44.3          & --      & --      & --      \\
 \rowcolor{gray!20}
ColaVLA & -- & \textbf{3.48}  & \textbf{3.54} & \textbf{3.16} & \textbf{3.75} & \textbf{36.8} & \textbf{32.3} & \textbf{51.6} & \textbf{26.4} \\

\bottomrule[1.5pt]
\end{tabular}
}
\caption{Closed-loop simulation results on NeuroNCAP~\cite{ljungbergh2024neuroncap}. $\dagger$ indicates that it utilizes additional training data and is a text-based driving VLM model. $\ddagger$ refers to trajectory post-processing. Our proposed method achieves a substantial improvement in the closed-loop evaluation, demonstrating strong adaptability to safety-critical scenarios and highlighting the model's efficiency and generalization capability. In this evaluation, we adopt only the \textbf{top-1 driving strategy} to better simulate realistic decision-making in closed-loop settings, ensuring fair comparison and faithfully reflecting the model's safety and robustness in real-world driving situations. 
Best scores are in \textbf{bold}.}

\label{table:NeuroNCAP}
\end{table*}

\subsection{Experiment Settings}

\noindent
\textbf{Datasets and evaluation metrics.}
We conduct experiments on the challenging nuScenes~\cite{caesar2020nuscenes} dataset, which contains 1,000 driving scenes of approximately 20 seconds each, providing six camera images, LiDAR data, semantic maps, and 3D bounding box annotations for keyframes. 
To enhance the planning-related supervision of our VLM, we additionally use the OmniDrive-nuScenes~\cite{wang2024omnidrive} extension, which augments nuScenes with QA-style annotations spanning perception, prediction, and planning. 
These annotations enable both offline reasoning supervision and online spatial reasoning.

We evaluate our model in both \textbf{open-loop} and \textbf{closed-loop} settings. 
Open-loop evaluation follows prior works~\cite{hu2023planning,jiang2023vad} and measures trajectory accuracy and safety using two standard metrics: the L2 displacement error (distance between predicted and ground-truth trajectories) and the collision rate. 
For closed-loop evaluation, we employ the NeuroNCAP simulator~\cite{ljungbergh2024neuroncap}, a photorealistic framework built upon nuScenes that reconstructs diverse, safety-critical urban interactions for realistic policy assessment. 
Performance is measured by the NeuroNCAP Score, a five-star metric that rewards safe, low-impact behavior, and the collision rate, jointly reflecting both safety and operational efficiency. 
Together, these benchmarks provide a comprehensive evaluation of the model's accuracy, robustness, and generalization in both static and dynamic driving scenarios.

\noindent
\textbf{Implementation details.} Our implementation is based on the LLaVA v1.5~\cite{liu2024visual} framework, which uses LLaMA-7B as the language model. 
Following prior work, we initialize the image encoder with EVA-02-L~\cite{fang2024eva} and adopt the SQ-Former~\cite{Chen_2025_CVPR} architecture for better visual reasoning. 
The detection decoder and lane decoder follow StreamPETR~\cite{wang2023exploring}, with 900 object queries and 300 lane queries, respectively.

We use a multi-stage training strategy. 
In the first stage, our VLM is pretrained on QA pairs from OmniDrive-nuScenes, where LoRA layers~\cite{hu2021lora} are applied to prompt and adapt the VLM efficiently for perception–planning alignment. 
In the second stage, we integrate our action-based planner and perform joint fine-tuning. Within the VLM, we only update the LoRA parameters to retain pretrained knowledge while improving decision capability. 
Training is conducted with the AdamW optimizer~\cite{adamw} with Cosine Annealing~\cite{cosineanneal}, using a weight decay of $1\times10^{-4}$ and an initial learning rate of $1\times10^{-4}$. Additional details and hyperparameters are provided in the supplementary material.

\subsection{Comparison with State-of-the-art Methods}

\noindent\textbf{Open-loop planning results.}
\cref{tab:sota-plan} reports results on the nuScenes open-loop benchmark. 
Among action-based approaches, ColaVLA attains the best overall accuracy and safety, achieving the lowest average L2 error ($0.30$m) and the lowest average collision rate ($0.23\%$).
Compared with the strongest prior action-based baseline, SOLVE-E2E (Avg.\ L2 $0.31$m; Avg. Col. $0.30\%$), our method reduces L2 by $3\%$ and collision rate by $23\%$, indicating more precise and safer trajectory predictions.
Notably, ColaVLA is also competitive with the latest text-based VLM planners while avoiding autoregressive text decoding. 
By relocating reasoning to a latent space and introducing cognitive latent reasoning with hierarchical parallel decoding, our framework reduces the number of VLM forward passes by over \textbf{5$\times$} fewer than typical text-based pipelines, highlighting its superior efficiency while operating directly in the latent action space.

\noindent\textbf{Closed-loop planning results.}
On the NeuroNCAP closed-loop benchmark \cref{table:NeuroNCAP}), ColaVLA establishes a new state of the art with a NeuroNCAP score of $3.16$, surpassing the strongest prior method, ImpromptuVLA, by an absolute margin of $+1.10$ ($53\%$ relative).
In terms of safety, our model lowers the average collision rate from $65.1\%$ to $42.5\%$, with particularly large gains on static collisions ($54.8\%\!\to\!15.0\%$; about $73\%$ reduction) and the best performance on side collisions. The overall lower collision profile and the substantially higher NeuroNCAP score highlight strong closed-loop robustness.
Importantly, ImpromptuVLA is a text-based VLM trained with additional curated data, whereas ColaVLA reaches higher scores without textual chain-of-thought inference and additional safety-critical data.
These results validate the effectiveness of our cognitive latent reasoning and hierarchical parallel planner: relocating reasoning to a vision-aligned latent space and decoding trajectories in a single parallel pass translates into better decision quality and improved safety under closed-loop evaluation.

\noindent\textbf{Inference latency comparison.}
As shown in \cref{tab:latency}, our ColaVLA achieves the lowest latency among all compared methods while maintaining strong planning accuracy and safety performance. Compared with SOLVE-VLM~\cite{Chen_2025_CVPR} and OmniDrive~\cite{wang2024omnidrive}, which depend on autoregressive chain-of-thought reasoning at the text level, our latent reasoning and single-pass hierarchical decoding deliver over \textbf{5$\times$} faster inference, enabling efficient and interpretable planning.

\begin{table}[!t]
\centering
\scalebox{1}{
\begin{tabular}{c|c|c}
\toprule[1.5pt]
Method & Action-based & Latency (ms) $\downarrow$ \\
\hline
OmniDrive~\cite{wang2024omnidrive} & \ding{55} & 3727 \\
SOLVE-VLM~\cite{Chen_2025_CVPR} & \ding{55} & 3719 \\
Ours & \ding{51} & \textbf{727} \\
\bottomrule[1.5pt]
\end{tabular}}
\vspace{-2mm}
\caption{
Inference latency comparison on a single NVIDIA H20 GPU without flash-attention~\cite{dao2022flashattention}.
We report end-to-end inference latency (ms per frame) under identical batch settings.}
\label{tab:latency}
\vspace{-2mm}
\end{table}

\subsection{Qualitative Results}
\begin{figure*}[t]
    \centering
    \vspace{-1mm}
    \includegraphics[width=\linewidth]{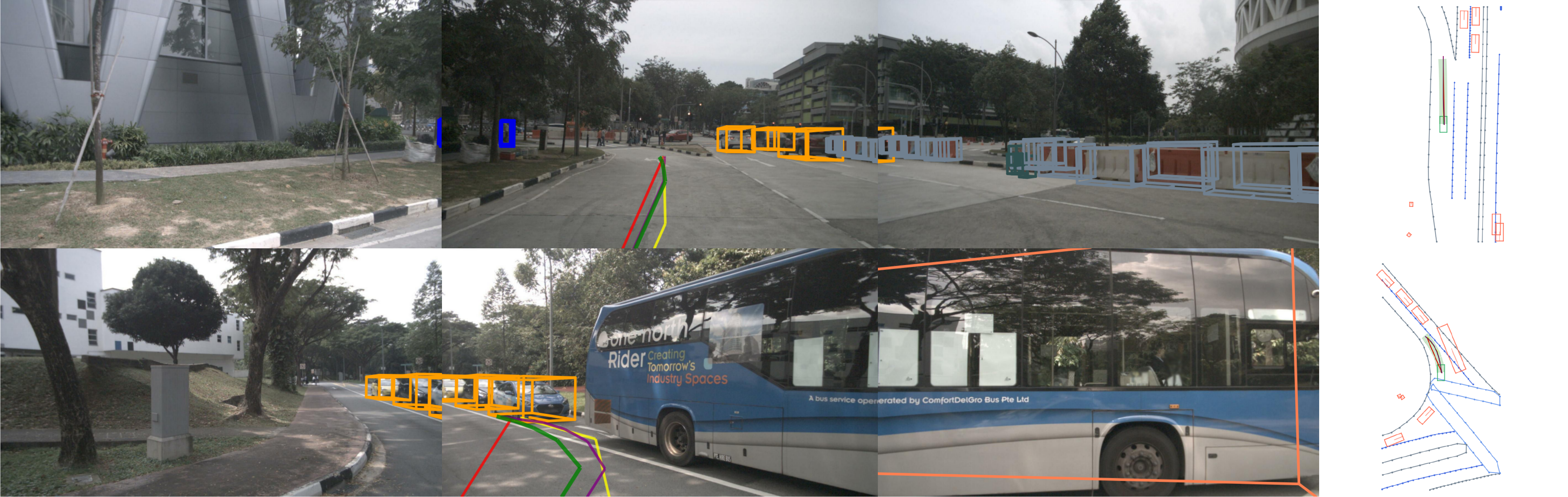}
    \caption{
    \textbf{Qualitative visualization of multi-scale trajectory predictions.}
    Red, yellow, and purple curves denote endpoint-only to full-trajectory predictions, while the green curve is the ground-truth. Right: BEV visualization with ego vehicle, agents, and trajectories.
    }
    \label{fig:visualize}
    \vspace{-2mm}
\end{figure*}

\cref{fig:visualize} shows qualitative results from our Hierarchical Parallel Planner. 
In both straight and turning scenarios, coarse trajectories (red) capture global intent, while finer scales (yellow, purple) progressively refine spatial details and curvature, converging closely to the ground truth (green). 
These results demonstrate that our hierarchical decoding produces smooth, and accurate plans within a single forward pass.

\subsection{Ablation Studies}
In this section, we systematically ablate the proposed components and key hyperparameters to assess their individual and combined contributions. We report results on nuScenes~\cite{caesar2020nuscenes} using two complementary metrics: the open-loop L2 average error and the closed-loop NeuroNCAP Score.

\noindent
\textbf{Ablation on Latent Reasoning.} As shown in \cref{tab:reasoning}, we conduct ablation experiments on our reasoning module to verify its effectiveness. 
Introducing latent reasoning significantly enhances the model's reasoning ability, enabling accurate prediction and reducing the average L2 error. 
Furthermore, adding the Rethink stage allows our model to re-evaluate the compressed key information from the current driving scene, refining visual understanding and facilitating better decision-making in subsequent stages. 
This progressive reasoning process improves the model's generalization and robustness, especially in complex or dynamic traffic scenarios.

\begin{table}[!t]
\centering
\scalebox{1}{
\begin{tabular}{c|c|cccc}
\toprule[1.5pt]
\multirow{2}{*}{Reasoning} & \multirow{2}{*}{Rethink} & \multicolumn{4}{c}{L2 (cm) $\downarrow$} \\

 &  & 1s & 2s & 3s & \cellcolor{gray!30} Avg \\
\hline
\ding{55} & \ding{55} & 14.1 & 28.5 & 54.2 & \cellcolor{gray!30} 32.2 \\
\ding{51} & \ding{55} & 14.5 & 28.5 & 50.7 & \cellcolor{gray!30} 31.3 \\
\ding{51} & \ding{51} & \textbf{14.0} & \textbf{27.1} & \textbf{50.2} & \cellcolor{gray!30} \textbf{30.4} \\
\bottomrule[1.5pt]
\end{tabular}}
\vspace{-2mm}
\caption{Ablation on Latent Reasoning. We evaluate the effect of the reasoning process for latent driving strategy inference and the reflective rethinking of critical cues within the cognitive reasoner.}
\label{tab:reasoning}
\vspace{-2mm}
\end{table}

\begin{table}[!t]
\centering
\scalebox{1}{
\begin{tabular}{c|cccc}
\toprule[1.5pt]
\multirow{2}{*}{Planner} & \multicolumn{4}{c}{NeuroNCAP Score $\uparrow$} \\
 & Static & Frontal & Side & \cellcolor{gray!30} Avg \\
\hline
MLP-based & 1.18 & 0.65 & 1.31 & \cellcolor{gray!30} 1.05 \\
Diffusion-based & 1.05 & 0.58 & \textbf{1.43} & \cellcolor{gray!30} 1.02 \\
Ours & \textbf{2.58} & \textbf{1.10} & 0.82 & \cellcolor{gray!30} \textbf{1.50} \\
\bottomrule[1.5pt]
\end{tabular}}
\vspace{-2mm}
\caption{Ablation on the action-based planner under closed-loop evaluation on NeuroNCAP benchmark. We use deterministic MLP and  stochastic diffusion heads to compare against our planner.}
\label{tab:planner_ablation_closedloop}
\vspace{-2mm}
\end{table}

\begin{table}[!t]
\centering
\scalebox{0.9}{
\begin{tabular}{c|cccc}
\toprule[1.5pt]
\multirow{2}{*}{Retained token number $K$} & \multicolumn{4}{c}{L2 (cm) $\downarrow$} \\

 & 1s & 2s & 3s & \cellcolor{gray!30} Avg \\
\hline
128  & 14.4 & 28.3 & 50.8 & \cellcolor{gray!30} 31.2 \\
192  & 14.2 & 28.1 & 50.5 & \cellcolor{gray!30} 30.9 \\
256 & \textbf{14.0} & \textbf{27.1} & \textbf{50.2} & \cellcolor{gray!30} \textbf{30.4} \\
320 & 14.6 & 28.8 & 51.8 & \cellcolor{gray!30} 31.7 \\
\bottomrule[1.5pt]
\end{tabular}}
\vspace{-2mm}
\caption{Ablation on the number of retained critical tokens $K$. }
\label{tab:critical_token}
\vspace{-2mm}
\end{table}

\begin{table}[!t]
\centering
\scalebox{1}{
\begin{tabular}{c|cccc}
\toprule[1.5pt]
\multirow{2}{*}{Strategy Type} & \multicolumn{4}{c}{L2 (cm) $\downarrow$} \\

 & 1s & 2s & 3s & \cellcolor{gray!30} Avg \\
\hline
Single scale & \textbf{13.8} & 28.4 & 53.4 & \cellcolor{gray!30} 31.9 \\
Multi-scale (Sequential) & 14.5 & 28.7 & 52.2 & \cellcolor{gray!30} 31.8 \\
Multi-scale (Reverse) & 14.7 & 28.5 & 51.1 & \cellcolor{gray!30} 31.4 \\
Multi-scale (Interpolate) & 14.0 & \textbf{27.1} & \textbf{50.2} & \cellcolor{gray!30} \textbf{30.4} \\
\bottomrule[1.5pt]
\end{tabular}}
\vspace{-2mm}
\caption{Ablation on the strategy of hierarchical regression. All variants share the same parallel decoding framework but differ in their specific selection strategy of trajectory subsets across scales.}
\label{tab:hierarchical_order}
\vspace{-4mm}
\end{table}

\noindent
\textbf{Ablation on action-based planner.}
As action-based planners can often achieve similar open-loop metrics, we focus here on their closed-loop performance to better assess real-world driving behavior. 
To isolate the planner's contribution, the reasoning module is disabled in this comparison. 
As shown in \cref{tab:planner_ablation_closedloop}, our Hierarchical Parallel Planner significantly outperforms both MLP- and diffusion-based planners under the NeuroNCAP benchmark. 
In particular, our method yields substantial improvements in static and frontal scenarios, highlighting its ability to generate smoother, more stable, and safer trajectories. 
By decoding trajectories from coarse to fine while preserving causal dependencies across temporal scales, our planner effectively refines intermediate predictions and adapts robustly to complex or dynamic traffic conditions, demonstrating superior safety and generalization.

\noindent
\textbf{Ablation on the number of retained critical tokens $K$.}
We further analyze the influence of the number of retained visual tokens $K$ in the ego-adaptive router. 
The original object and lane queries contain $900$ and $300$ tokens, respectively. 
As shown in \cref{tab:critical_token}, retaining too few tokens limits the representational capacity, causing significant loss of visual information and degraded performance. 
Conversely, retaining too many tokens introduces redundancy, increasing computational overhead during both training and inference. 
Finally we choose $K{=}256$ as our default configuration, which achieves the best trade-off, providing sufficient semantic coverage while maintaining computational efficiency.

\noindent
\textbf{Ablation on the strategy of hierarchical regression.}
We investigate different strategies for hierarchical trajectory regression, all under the same parallel decoding framework but with varied ordering schemes. 
The Single scale baseline directly regresses the final trajectory without temporal abstraction, serving as a non-causal reference. 
The Sequential strategy extends trajectories forward from the start, while the Reverse strategy begins from the final point and propagates backward. 
Our proposed Interpolate strategy instead first predicts key endpoints and then fills in intermediate points across scales, aligned with the causal structure of driving motion. 
As shown in \cref{tab:hierarchical_order}, all multi-scale designs outperform the single-scale baseline, demonstrating the benefits of temporal abstraction. 
Among them, the Interpolate strategy yields the best performance, validating its effectiveness in structured and causally consistent trajectory reasoning.

\section{Conclusion}
We introduced ColaVLA, a unified vision–language–action framework for end-to-end autonomous driving. By relocating reasoning from text to a unified latent space and coupling it with a hierarchical parallel planner, ColaVLA bridges the gap between VLM cognition and continuous action generation. 
Its cognitive latent reasoner efficiently compresses scene understanding into compact meta-action representations through ego-adaptive token selection, while the causality-preserving planner decodes multi-scale trajectories in a single forward pass. 
This design achieves efficient, interpretable, and safe planning with minimal latency.
Experiments on nuScenes demonstrate that ColaVLA sets new state-of-the-art performance in both open-loop and closed-loop evaluations while maintaining strong generalization, efficiency, and robustness. 
These results suggest that transferring reasoning from text to latent space provides a scalable path toward efficient, knowledge-driven autonomous driving systems.

{
    \small
    \bibliographystyle{ieeenat_fullname}
    \bibliography{main}
}

\clearpage
\setcounter{page}{1}
\maketitlesupplementary

\section{Further Ablation Study}
\label{sec:suppl-ablation}

\noindent
\textbf{Ablation on context tokens for planning.}
We further study whether the planner should attend to all visual tokens or only to the pruned subset selected by the ego-adaptive router. 
Concretely, we compare two variants: one that uses the full set of original vision tokens when predicting the final trajectory, and one that relies solely on the pruned critical tokens $\mathbf{Q}^*$ while keeping all other components unchanged.
As summarized in \cref{tab:prune_token}, using pruned tokens leads to lower L2 errors compared to using all tokens. 
This indicates that the ego-adaptive router can accurately identify which visual cues are most relevant to the current driving scenario and future decisions, effectively filtering out redundant background information. 
At the same time, pruning substantially reduces the token length fed into the planner, thereby lowering computation and improving inference efficiency without sacrificing final prediction accuracy.

\begin{table}[!t]
\centering
\scalebox{1}{
\begin{tabular}{c|cccc}
\toprule[1.5pt]
\multirow{2}{*}{Visual tokens} & \multicolumn{4}{c}{L2 (cm) $\downarrow$} \\
 & 1s & 2s & 3s & \cellcolor{gray!30} Avg \\
\hline
Full tokens & 14.8 & 28.9 & 52.4 & \cellcolor{gray!30} 32.0 \\
Pruned tokens (Ours) & \textbf{14.0} & \textbf{27.1} & \textbf{50.2} & \cellcolor{gray!30} \textbf{30.4} \\
\bottomrule[1.5pt]
\end{tabular}}
\vspace{-2mm}
\caption{Ablation on planner context token selection. Using the pruned critical tokens as context for our planner yields better accuracy while reducing sequence length and computational cost.}
\label{tab:prune_token}
\vspace{-2mm}
\end{table}

\section{Further Implementation Details}
\label{sec:suppl-implement}

\noindent
\textbf{Architecture.}
Our implementation is built on the LLaVA v1.5 framework~\cite{liu2024visual}, which uses LLaMA-7B as the language backbone and introduces LoRA adapters~\cite{hu2021lora} for efficient adaptation. 
The image encoder follows EVA-02-L~\cite{fang2024eva} instantiated as an EVAViT backbone with 24 transformer layers, hidden dimension 1024 and window size 16, initialized from publicly available weights. 
On top of the image backbone, we adopt the SQ-Former architecture~\cite{Chen_2025_CVPR} for multi-view visual reasoning and use perception task heads.
We use 900 object queries and 300 lane queries by default, consistent with our main experiments. 
The overall system is implemented in a single camera-only configuration with 6 multi-view images per frame. And point-cloud-style BEV parameters (e.g., point cloud range and voxel size) are used only for defining the spatial grid for 3D perception heads.

\noindent
\textbf{Multi-stage training strategy.}
We employ a three-stage training pipeline to fully exploit the strong VLM priors while stabilizing downstream hierarchical planning:

\begin{itemize}[leftmargin=1.4em,itemsep=2pt]
  \item \emph{Stage 1: VLM adaptation on OmniDrive QA.}  
  We first adapt the vision–language backbone on OmniDrive-nuScenes QA pairs~\cite{wang2024omnidrive}. 
  Similar to OmniDrive, we use scene-level QA covering description, perception, prediction and planning questions, and optimize only the LoRA layers on top of the LLaMA-7B language model together with the Q-Former style visual interface. 
  This stage encourages the VLM to encode driving-relevant semantics and reasoning skills in complex urban scenes without changing the base model weights, and aligns the visual encoder with driving-oriented questions.

  \item \emph{Stage 2: Meta-action bank pretraining.}  
  In the second stage, we introduce the meta-action bank used by the Cognitive Latent Reasoner. 
  We cluster nuScenes trajectories (using the provided $k$-means anchors in the config) to obtain a discrete set of meta-action prototypes, which are used to initialize the meta-action queries and the action bank. 
  During this stage, we freeze the language backbone VLM (except LoRA), and primarily train the meta-action classification and trajectory prediction heads to associate latent meta-actions with typical driving patterns (e.g., straight, lane change, turn, braking). 
  This provides a structured and stable prior for downstream hierarchical planning.

  \item \emph{Stage 3: End-to-end fine-tuning on nuScenes.}  
  Finally, we fine-tune ColaVLA on nuScenes planning supervision. 
  The language backbone remains frozen while LoRA adapters, the visual encoder (EVAViT + SQ-Former), detection/map heads, cognitive latent reasoning components (router and meta-action queries), and the hierarchical parallel planner are all jointly optimized. 
  We train with multi-scale trajectory losses and the top-$K$ hypothesis strategy, which enables confidence-guided selection among multiple meta-action hypotheses.
\end{itemize}

\noindent
\textbf{Optimization and training schedule.}
Unless otherwise specified, all stages are trained on 16 NVIDIA H20 GPUs with a per-GPU batch size of 2 (effective batch size 32). 
We use the AdamW optimizer~\cite{adamw} with initial learning rate $1\times10^{-4}$, weight decay $1\times10^{-4}$, and $(\beta_1,\beta_2)=(0.9,0.999)$. 
The learning rate is scheduled by Cosine Annealing~\cite{cosineanneal} with linear warmup for the first 500 iterations and a minimum learning-rate ratio of $10^{-3}$. 
We train for 10 epochs on nuScenes dataset in the final stage, using mixed-precision (FP16) with dynamic loss scaling and gradient clipping at a global norm of 10. 

\noindent
\textbf{Data processing.}
We follow the standard nuScenes setting~\cite{caesar2020nuscenes} with the official train/val split. 
Our data loader uses \texttt{CustomNuScenesDataset} with temporal sequences, camera-only inputs, and the same object classes as the main experiments. 
Images are first loaded at $900\times1600$ resolution and then augmented with random resize–crop–rotation as in the provided augmentation configure. 
For training, we resize each view to $320\times640$ and normalize them using ImageNet statistics. 
We adopt the same pipeline for validation but disable random augmentation. 
Planning supervision is derived from nuScenes ego trajectories, while QA supervision and lane-object relations are loaded from the OmniDrive and preprocessed lane-object files.

\noindent
\textbf{Summary of key hyperparameters.}
For clarity, \cref{tab:impl_hparams} summarizes the main architectural and training hyperparameters used in our implementation.

\begin{table*}[t]
\centering
\small
\begin{tabular}{l|p{0.7\textwidth}}
\toprule[1.2pt]
\textbf{Category} & \textbf{Setting} \\
\midrule
Backbone VLM & LLaVA v1.5 (LLaMA-7B) with LoRA adapters \\
Image encoder & EVA-02-L (EVAViT, 24 layers, 1024-d, window size 16) \\
Visual reasoning & SQ-Former with temporal PETR-style transformers \\
Object queries & 900 queries (3D detection head) \\
Lane queries & 300 queries (map / lane head) \\
Input modality & 6-camera views, camera-only, no LiDAR or radar \\
Image resolution & $900\times1600 \rightarrow 320\times640$ after augmentation \\
Point cloud range & $[-51.2, -51.2, -5.0,\; 51.2, 51.2, 3.0]$ \\
Voxel size & $[0.2, 0.2, 8.0]$ (for BEV grid definition) \\
Planner stages & $S=6$ hierarchical scales (as in \texttt{stage\_index}) \\
Multi-scale loss & Weights $[0.5, 0.7, 1.0, 1.2, 1.5, 1.8]$ \\
Top-$K$ hypotheses & $K=3$ meta-action modes (\texttt{topk\_mode\_predict=3}) \\
Batch size & 2 samples / GPU ($\times 16$ GPUs) \\
Epochs (stage 3) & 10 epochs on nuScenes \\
Optimizer & AdamW, lr $1\times 10^{-4}$, wd $1\times 10^{-4}$, $(0.9,0.999)$ \\
LR schedule & Cosine Annealing + 500-iter linear warmup, min lr ratio $10^{-3}$ \\
Precision & FP16 with dynamic loss scaling \\
Grad clipping & Global norm 10 \\
Efficiency tricks & Gradient checkpointing, FlashAttention, mixed precision \\
\bottomrule[1.2pt]
\end{tabular}
\caption{Main implementation and training hyperparameters for ColaVLA.}
\label{tab:impl_hparams}
\end{table*}

\begin{figure*}[t]
    \centering
    \vspace{-1mm}
    \includegraphics[width=\linewidth]{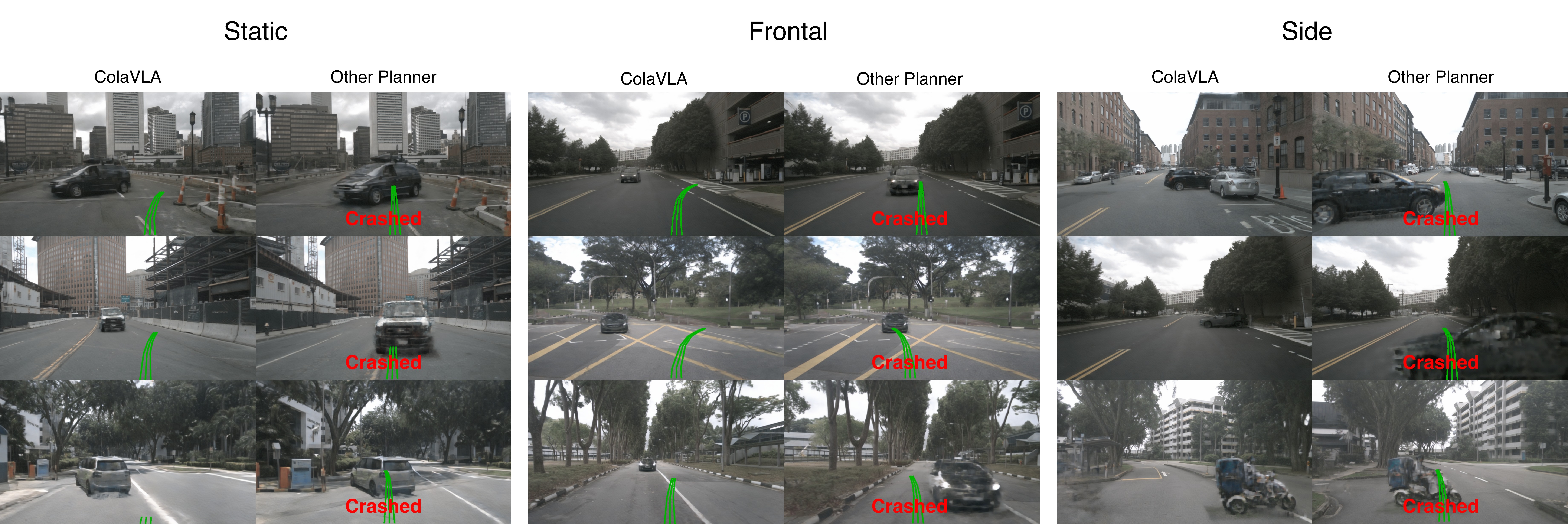}
    \caption{
    Qualitative closed-loop comparisons in the NeuroNCAP simulator across three representative scenario types. For each case, we visualize the predicted trajectories of ColaVLA and competing planners. ColaVLA consistently guides the ego vehicle away from potential collisions, producing safer and more stable motions.
    }
    \label{fig:suppl_neuroncap}
    \vspace{-2mm}
\end{figure*}

\section{Additional Closed-Loop Visualizations}
\label{sec:suppl-qualitative}

We further provide qualitative results from the NeuroNCAP simulator to complement the nuScenes open-loop visualizations in the main paper. As illustrated in \cref{fig:suppl_neuroncap}, our model produces stable and collision-averse behaviors across diverse safety-critical scenarios, including static obstacles, frontal interactions, and side conflicts. Compared with previous text-based VLM planners, ColaVLA exhibits more responsive closed-loop reactions and fewer failure cases, while simultaneously maintaining higher inference efficiency. These examples corroborate our quantitative findings that ColaVLA achieves stronger NeuroNCAP scores and lower latency, delivering both safer and faster closed-loop planning.

\section{Details about Meta-Action Bank}
\label{sec:suppl-meta-action}

To support cognitive latent reasoning, we pre-compute a discrete meta-action label for each nuScenes sample and use it to index the action bank. Each meta-action is defined by combining a \emph{path pattern} with a \emph{speed pattern} derived jointly from the future ego trajectory and CAN bus signals. In our final setup, this yields eight meta-action classes: four straight-driving modes (stopping, constant-speed cruising, accelerating, and decelerating along a roughly straight path) and four turning or lane-change modes (large left turns or U-turns, large right turns or U-turns, small left turns or lane changes, and small right turns or lane changes). Our implementation broadly follows the trajectory categorization procedure used in Senna~\cite{jiang2024senna}.

\noindent\textbf{Geometry and CAN bus fusion.}
For each sample, we first obtain a future ego trajectory in the ego frame as a sequence of $T$ waypoints at a fixed time interval. If the planning trajectory is missing or fully padded, we reconstruct it from nuScenes CAN bus pose messages by transforming global positions into the initial ego frame and resampling to $T$ steps; if this is still not possible, we fall back to a zero trajectory. In parallel, we read CAN bus logs around the planning horizon (pose, yaw, longitudinal velocity and acceleration), and compute fused heading and speed statistics such as cumulative yaw angle, maximum yaw rate, final speed, speed change, and mean acceleration. When available, CAN bus statistics are preferred; otherwise, we rely on geometric estimates from the trajectory itself.

\noindent\textbf{Path classification.}
We first assign each sample to one of five path types: straight motion, large left turn (including left U-turns), large right turn (including right U-turns), small left turn or lane change, and small right turn or lane change. The assignment is determined using fused heading and lateral-motion statistics. Large turns are identified by large cumulative yaw with a clear left or right sign; small turns correspond to moderate yaw changes; lane-change-like motions are characterized by small net yaw but significant lateral displacement with consistent lateral direction and bounded yaw rate; straight motion is defined by small yaw and small lateral offset. Ambiguous residual cases are assigned to left or right small turns according to the sign of the lateral displacement. This step yields a path label for every sample.

\noindent\textbf{Speed classification and heavy paths.}
Speed patterns are only refined for path types that appear frequently in the training set. We first count occurrences of each path type and mark \emph{heavy} paths whose counts exceed a global threshold. In practice, only straight paths are heavy, while turning and lane-change paths are relatively sparse.

For heavy paths, we further classify the speed profile into four modes: stopping, approximately constant speed, accelerating, and decelerating. The decision uses fused speed statistics: very low final speed indicates stopping; clearly increasing speed or positive mean acceleration indicates acceleration; clearly decreasing speed or negative mean acceleration indicates deceleration; otherwise the sample is treated as constant-speed. For non-heavy path types, we do not split by speed and simply mark the speed pattern as “any”, in order to avoid severe class imbalance.

\noindent\textbf{Final meta-action labels.}
The final meta-action class for each sample is defined as the combination of its path type and speed type. After applying the heavy-path and speed rules above, only eight combinations are active in training: four straight-driving modes (stop, constant-speed, accelerate, decelerate along a straight path) and four turning or lane-change modes (large left, large right, small left, small right). Each combination is assigned a stable integer id, which we precompute offline and store alongside each nuScenes sample. During training, the cognitive latent reasoner predicts a distribution over these eight meta-actions, and the top-scoring modes are used to retrieve corresponding action priors from the action bank, providing structured, trajectory-aligned guidance for the hierarchical parallel planner.

\end{document}